\documentclass[letterpaper, 10 pt, conference]{ieeeconf}  % Comment this line out if you need a4paper
\IEEEoverridecommandlockouts          % This command is only needed if
\usepackage{amsmath,amsfonts}
\usepackage{algorithmic}
\usepackage{array}
\usepackage[caption=false,font=normalsize,labelfont=sf,textfont=sf]{subfig}
\usepackage{textcomp}
\usepackage{stfloats}
\usepackage{url}
\usepackage{verbatim}
\usepackage{graphicx}
\usepackage{multirow}
\usepackage{booktabs}
\usepackage{caption}
\usepackage{lipsum}
\usepackage{balance}
\usepackage{hyperref}
\usepackage{CJKutf8}
\usepackage{cite}
\usepackage{url}
\hypersetup{
	colorlinks=true,            % 激活链接颜色，去掉链接边框
	linkcolor=red,              % 文档内部链接颜色（如图表等引用）
	citecolor=green,            % 文献引用链接颜色
	filecolor=blue,   % 文件链接颜色
	urlcolor=blue            % 外部URL链接颜色
}
\begin{document}
\title{ModaLink: Unifying Modalities for Efficient Image-to-PointCloud \\ Place Recognition}
\author{Weidong Xie, Lun Luo, Nanfei Ye, Yi Ren, Shaoyi Du$^*$, Minhang Wang, \\ Jintao Xu, Rui Ai, Weihao Gu, Xieyuanli Chen$^*$ % <-this % stops a space
\thanks{ * indicates corresponding authors.}%
\thanks{W. Xie and S. Du are with National Key Laboratory of Human-Machine Hybrid Augmented Intelligence, National Engineering Research Center for Visual Information and Applications, and Institute of Artificial Intelligence and Robotics, Xi'an Jiaotong University, Xi'an 710049, China, \{xieweidongxjtu, dushaoyi\}@stu.xjtu.edu.cn.
W. Xie is also with Haomo.AI Technology Co., Ltd.	
L. Luo, M. Wang, J. Xu, R. Ai, W. Gu, N. Ye are with Haomo.AI Technology Co., Ltd. X. Chen is with National University of Defense Technology. Y. Ren is with Carnegie Mellon University.}%
\thanks{This work was supported by the National Key Research and Development Program of China under Grant No. 2020AAA0108100, the National Natural Science Foundation of China under Grant Nos. 62327808, 61971343 and 62088102, and the Clinical Research Center of Shaanxi Province for Dental and Maxillofacial Diseases, College of Stomatology, Xi'an Jiaotong University under Grant No. 2021YHJB04.
 }%
}

% \markboth{Journal of \LaTeX\ Class Files,~Vol.~18, No.~9, September~2020}%
% {How to Use the IEEEtran \LaTeX \ Templates}

\maketitle

\begin{abstract}
Place recognition is an important task for robots and autonomous cars to localize themselves and close loops in pre-built maps. While single-modal sensor-based methods have shown satisfactory performance, cross-modal place recognition that retrieving images from a point-cloud database remains a challenging problem. 
Current cross-modal methods transform images into 3D points using depth estimation for modality conversion, which are usually computationally intensive and need expensive labeled data for depth supervision. 
In this work, we introduce a fast and lightweight framework to encode images and point clouds into place-distinctive descriptors. We propose an effective Field of View (FoV) transformation module to convert point clouds into an analogous modality as images.
This module eliminates the necessity for depth estimation and helps subsequent modules achieve real-time performance. We further design a non-negative factorization-based encoder to extract mutually consistent semantic features between point clouds and images. This encoder yields more distinctive global descriptors for retrieval. Experimental results on the KITTI dataset show that our proposed methods achieve state-of-the-art performance while running in real time. Additional evaluation on the HAOMO dataset covering a 17 km trajectory further shows the practical generalization capabilities. We have released the implementation of our methods as open source at: \url{https://github.com/haomo-ai/ModaLink.git}.

\end{abstract}

% \begin{IEEEkeywords}
% Place Recognition, Deep Learning Methods, Cross-Modality, Global Descriptor, Loop Closure Detection.
% \end{IEEEkeywords}

\section{Introduction}
\begin{figure}[!t]
	%\vspace{-0.5em}   %调整图表上方与正文之间的距离
	%\setlength{\abovecaptionskip}{-0.1cm}   %调整图片标题与图距离   
	\setlength{\belowcaptionskip}{-0.2cm}
	\centering
	\includegraphics[width=0.5\textwidth]{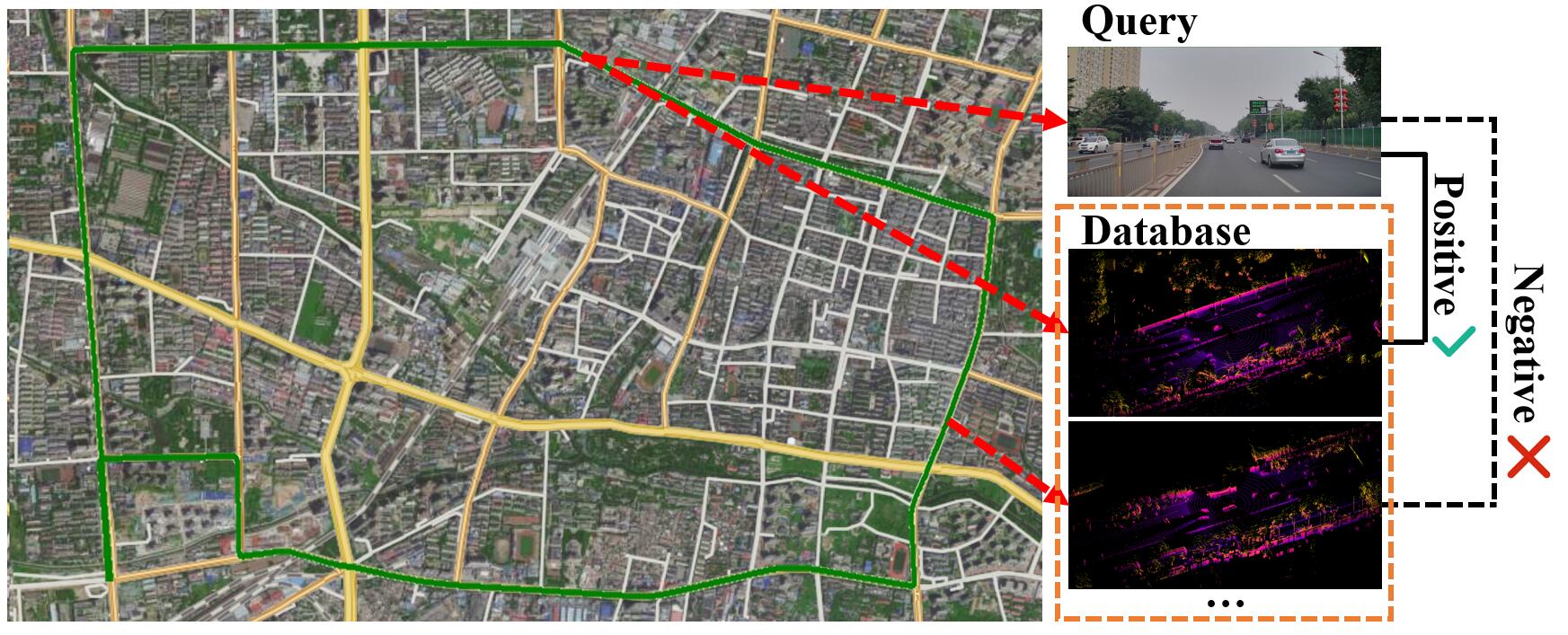}
	\caption{The purpose of ModaLink is to retrieve the most positive corresponding point cloud of the query image in a pre-built large-scale point cloud database.}
	\label{overallbaoding}
    \vspace{-1.0em}
\end{figure}

Place recognition is a key module of autonomous vehicle navigation systems. It locates sensor data, such as images and point clouds, on a pre-built map or database for global localization and loop closure detection. 
LiDAR-based place recognition utilizes precise structural information of point clouds to generate discriminative descriptors \cite{chen2020overlapnet, OverlapTransformer, ma2023cvtnet}. 
Since LiDAR sensors are expensive and bulky, deploying such methods is not cost-effective. As an alternative solution, visual place recognition is competitive due to its lower price and smaller size \cite{arandjelovic2016netvlad, zhu2022transgeo} of cameras. 
However, it is not robust to illumination changes and view variations \cite{zywanowski2020comparison}. 
Since different sensors have varying advantages, it can occur in real robotic applications that the data queried by one robot has different modalities compared to the database recorded by another, making cross-modal place recognition highly necessary. 
For instance, autonomous vehicles equipped with cheap cameras can use such methods to locate input images on large-scale point cloud databases and obtain accurate localization results, as shown in Fig. \ref{overallbaoding}. 
% ModaLink achieves global localization by finding the point cloud that is most similar to the image currently captured by the vehicle based on the point cloud map.

% Example of our image to point cloud place recognition method.

Generating accurate global descriptors that represent scene information is crucial for successful place recognition. However, designing these descriptors for cross-modal place recognition is very challenging, since data from different sensors such as LiDAR and camera are vastly different. Specifically, the LiDAR captures depth information by measuring the laser time of flight (TOF), while the camera collects photometric data by responding to the ambient light through CMOS. As both sensors capture different types of data, it is difficult to develop a unified feature extractor. To handle this issue, some methods~\cite{zheng2023i2p, lc2} transform different types of data into intermediate formats, such as bird's eye views (BEVs) and range images. They use depth estimation algorithms to transform images into 3D points and align them with the LiDAR point cloud. Although these methods perform well on public datasets, they cannot run in real time due to the high computational requirements of the depth estimation module.

In this work, we present a new approach that uses an FoV transformation module to bring together the data of images and point clouds. Instead of relying on depth estimation, our module converts 3D point clouds into 2D depth images. This makes it possible for subsequent modules to run in real-time, without performing time-consuming depth estimation. To further improve the distinctiveness of descriptors, we propose a non-negative factorization-based encoder to mine the implicit common features in an unsupervised way and use a shared-weight network to learn global descriptors based on the unified data formats. We thoroughly evaluated our ModaLink on both the KITTI and our self-collected dataset covering a 17\,km trajectory. The experimental results demonstrate that our method achieves state-of-the-art performance in cross-modal place recognition. 

In summary, our main contributions are as follows:
\begin{itemize}
    \item We propose a lightweight cross-modal place recognition method called ModaLink based on FoV transformation while using only monocular images. 
    \item We introduce a Non-Negative Matrix Factorization-based module to extract extra potential semantic features to improve the distinctiveness of descriptors.
    \item Extensive experimental results on the KITTI and a self-collected dataset show that our proposed method can achieve state-of-the-art performance while running in real-time at about 30Hz.
\end{itemize}
% \vspace{0.2cm}

% In the following part of this paper, we first review recent place recognition methods, then introduce the image-to-point cloud place recognition method based on FoV transformation, and detail the NMF-based encoder module. Finally, we evaluate the performance of the algorithm through a series of experiments.

\section{Related Work}

In this section, we first review the place recognition methods based on single-modal data. We then discuss the recent development of the approaches to achieve image-to-PointCloud place recognition.

%As a widely researched topic, existing works for image retrieval \cite{imageretrieval} and visual place recognition \cite{ipr4} systems have performed well on publicly available datasets.
\subsection{Visual Place Recognition (VPR)}
Early visual place recognition methods~\cite{2017ORB, mur2017orb} concentrate on developing hand-engineered features~\cite{rublee2011orb, lowe2004distinctive, 2012Bags}. These methods assume that similar environmental structures will result in comparable feature distributions. However, this assumption may not always apply for VPR, particularly when the appearances of environments vary across seasons. With the development of CNN architectures, CNN-based VPR methods have been developed. These methods transform the image features into localization descriptors, which improves the VPR performance by designing better local feature extractors and feature aggregators. For example, NetVLAD~\cite{arandjelovic2016netvlad} extracts local features by CNN and designs a network to learn the cluster centers. Patch-NetVLAD~\cite{hausler2021patchnetvlad} follows NetVLAD and leverages the strengths of both the global and local features to generate more distinct patch-level features. By learning from large datasets, these methods have exhibited better robustness and generalization ability in long-term~\cite{hausler2021patchnetvlad} and large-scale~\cite{zhu2018visual} tasks. More recently, transformers have been broadly used in the VPR problem. Yang et al.~\cite{yang2021cross} combine CNN and transformers and build a CNN+transformer structure. TransGeo~\cite{zhu2022transgeo} introduces the vision transformer~\cite{dosovitskiy2021image} structure and explicitly removes the invalid parts of the input images. With the help of transformers, these methods extract more distinct descriptors.
However, those methods focus only on VPR and cannot localize the visual descriptors on point cloud maps.

\subsection{LiDAR Place Recognition (LPR)}

In recent years, LiDAR-based place recognition has rapidly developed due to the ability of LiDAR to obtain accurate depth information and its robustness to changes in illumination and view.
Following the setup of the visual counterparts NetVLAD~\cite{arandjelovic2016netvlad}, PointNetVLAD~\cite{angelina2018pointnetvlad} utilizes the point-based encoder, i.e. PointNet~\cite{qi2017pointnet} for local feature extraction, and then generates global features with NetVLAD and achieve promising LPR results. 
However, it cannot capture local geometric structures due to its approach of treating each point independently. Thus, subsequent methods primarily focus on exploiting neighborhood information \cite{lpdnet, xia2021soe}. LiDAR point clouds are disordered and massive, thus hard to process using advanced vision learning techniques. Therefore, some methods project point clouds into images and encode the images into descriptors. Kim et al.~\cite{kim2018scan} propose the scan context descriptor by projecting point clouds into an ego-centric coordinate system. This system partitions the ground space into bins according to both azimuthal and radial directions. OverlapNet~\cite{chen2020overlapnet} and its later develops~\cite{chen2022overlapnet, OverlapTransformer, ma2023cvtnet} project 
point clouds into range images and adopt a siamese network to learn the overlap between a pair of images. BVMatch \cite{bvmatch2021} projects point clouds into BEV images and extracts the global BVFT feature for pose estimation. It shows that the projected image is a quite effective representation of point cloud and these methods \cite{bvmatch2021, luo2023bevplace} achieve state-of-the-art performance in terms of retrieval recall, robustness, and generalization ability.

% Traditional handcrafted methods \cite{tra1} and more learning-based methods \cite{bevplace} have been proposed to tackle the problems. Among CNN-based methods, projection-based methods \cite{overlapnet, overlaptrans} and point-based methods \cite{lpdnet, LCDNet} are two main parts.  \cite{PointNet} directly extracted features from geometrically local 3D points, consumes unordered 3D point sets as inputs, and is used in segmentation, classification, and LPR. \cite{pointnetvlad} first incorporate \cite{PointNet} into VLAD architecture. Point-based methods have made remarkable progress, but considerable computation consumption hinders their deployment in cars. Thus, many other works use 2D image-view to replace raw 3D point sets and focus on local-global descriptor extraction in 2D ways. \cite{overlapnet, overlaptrans} projected point cloud into range image view (RIV) and \cite{bevplace, bvmatch} transformed cloud into Bird-Eye View (BEV) image, extract local-global descriptors with 2D group convolution. \cite{NetVLAD} provided a baseline for the following point-based methods. \cite{cvtnet} exploited multi-layer both RIVs and BEVs of LiDAR data and employed unified transformer architecture to make yaw-invariant global descriptors.

\subsection{Image to Point Cloud Place Recognition}

% The cross-modal place recognition task based on image and point cloud data has been widely researched. These methods try to leverage 

% Considering the economic cost, LiDAR is not always available in low-budget civilian robots. Due to the lability of the camera to illumination and color changing, easily accessible image data is insufficient for the establishment of a cross-modal matching database. A feasible plan is to equip the map acquisition vehicle specifically with LiDARs and mass-produced robots with cameras. Then, we particularly focus on the image-to-pointcloud task. Here comes the question of how to locate images on a large-scale point cloud map. 

%convert the image into depth image with stereo matching methods and transform the depth image to a point cloud with intrinsic, which converts the cross-modal problem into LPR. For example, 

Although place recognition based on single-modal data has been extensively studied, identifying images on a point-cloud map remains a difficult task. As a result, there are fewer related works in comparison to the single-modal-based counterparts. 
One straightforward solution is to jointly train a 2D neural network for images and a 3D network for point clouds to create shared embeddings~\cite{cattaneo2020global}. However, this approach does not generalize well to unseen environments. Bernreiter et al.~\cite{bernreiter2021spherical} proposed to project images and point clouds into unit spheres and then extract features through such unit representations. This method requires multiple images as input, which is not always available. There are also approaches~\cite{feng20192d3d, chen2022i2d} detecting local features from images and point clouds to find cross-modal correspondences for place retrieval. 2D3D-MatchNet~\cite{feng20192d3d} designed a CNN network to cross-match the features in both the images and ground-removed point cloud submaps, while P2-Net~\cite{wang2021p2} selected a batch-hard detector to produce shared embedding between different sensor modalities. However, they perform only on a small scale and can hardly be extended to a large-scale environment well due to the heave point cloud representation. Another intuitive solution is to convert two types of data into the same modality. To this end, LC$^2$~\cite{lc2} exploited depth estimation to transform both images and point clouds into depth images. I2P-Rec~\cite{zheng2023i2p} adopted monocular and stereo depth estimation to convert 2D images into 3D point clouds and then utilized BEV-based methods to achieve LPR. However, the photometric rendering process is also computationally expensive.
The stereo matching and monocular depth estimation modules require labeled data for training and are usually of low running speed. 
Unlike the previous approaches, we design a system to achieve place recognition without depth estimation modules.

\section{Image-to-pointcloud Place Recognition based on FoV Transformation}

\begin{figure*}[!t]
	%\vspace{-1em}   %调整图表上方与正文之间的距离
	\setlength{\abovecaptionskip}{-0.1cm}   %调整图片标题与图距离   
	\setlength{\belowcaptionskip}{-0.3cm}
	\centering
	\includegraphics[width=\textwidth]{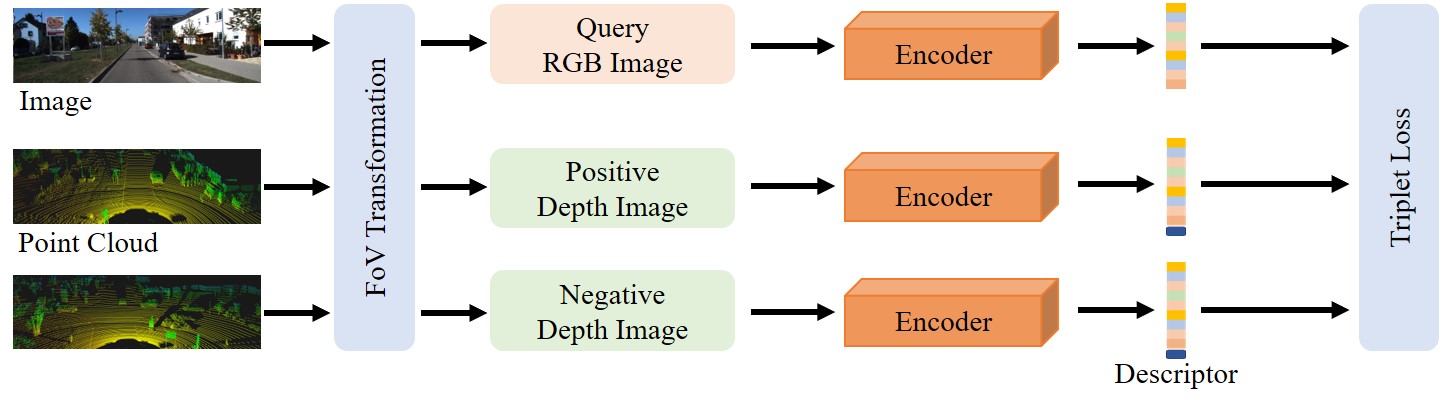}
	\caption{The training framework of our proposed ModaLink framework. Point clouds are converted to depth images via projection. Then, the query image and depth images are cropped into the same overlap of FoV. Based on the depth completion module, sparse depth images are upsampled to dense depth images. Then, global place descriptors are generated by a shared-weight encoder. Finally, we adopt triplet loss for supervision.}
	\label{schema}
\end{figure*}

The goal of this paper is to develop a method for fast and accurate image-to-pointcloud place recognition. Specifically, it can be defined as: given an image, the task is to retrieve the corresponding point cloud from the database that recorded at the same place. To this end, we propose ModaLink based on a novel FoV transformation module as shown in Fig.~\ref{schema}. For a query image, we find its positive and negative point clouds from the training set. Then, we use an FoV transformation module to unify the data format. We use a shared-weight encoder based on NMF to generate global descriptors. Finally, triplet loss is adopted for supervision. We elaborate on our designs next.

\subsection{FoV Transformation}

The objective of the FoV transformation module is to unify images and point clouds into a similar data format. As illustrated in Fig.~\ref{align}, the FoV transformation module first aligns images and point clouds by projecting the point clouds into depth images with the extrinsic and intrinsic parameters. It then crops both depth and RGB images into the same FoV. After obtaining the sparse depth images, it completes the sparse depth image into a denser one for more distinctive feature extraction. More details for each step are as follows:

\begin{figure}[!t]
	%\vspace{-0.5em}   %调整图表上方与正文之间的距离
	%\setlength{\abovecaptionskip}{-0.1cm}   %调整图片标题与图距离    
	%\setlength{\belowcaptionskip}{-0.5cm}
	\centering
	\includegraphics[width=0.48\textwidth]{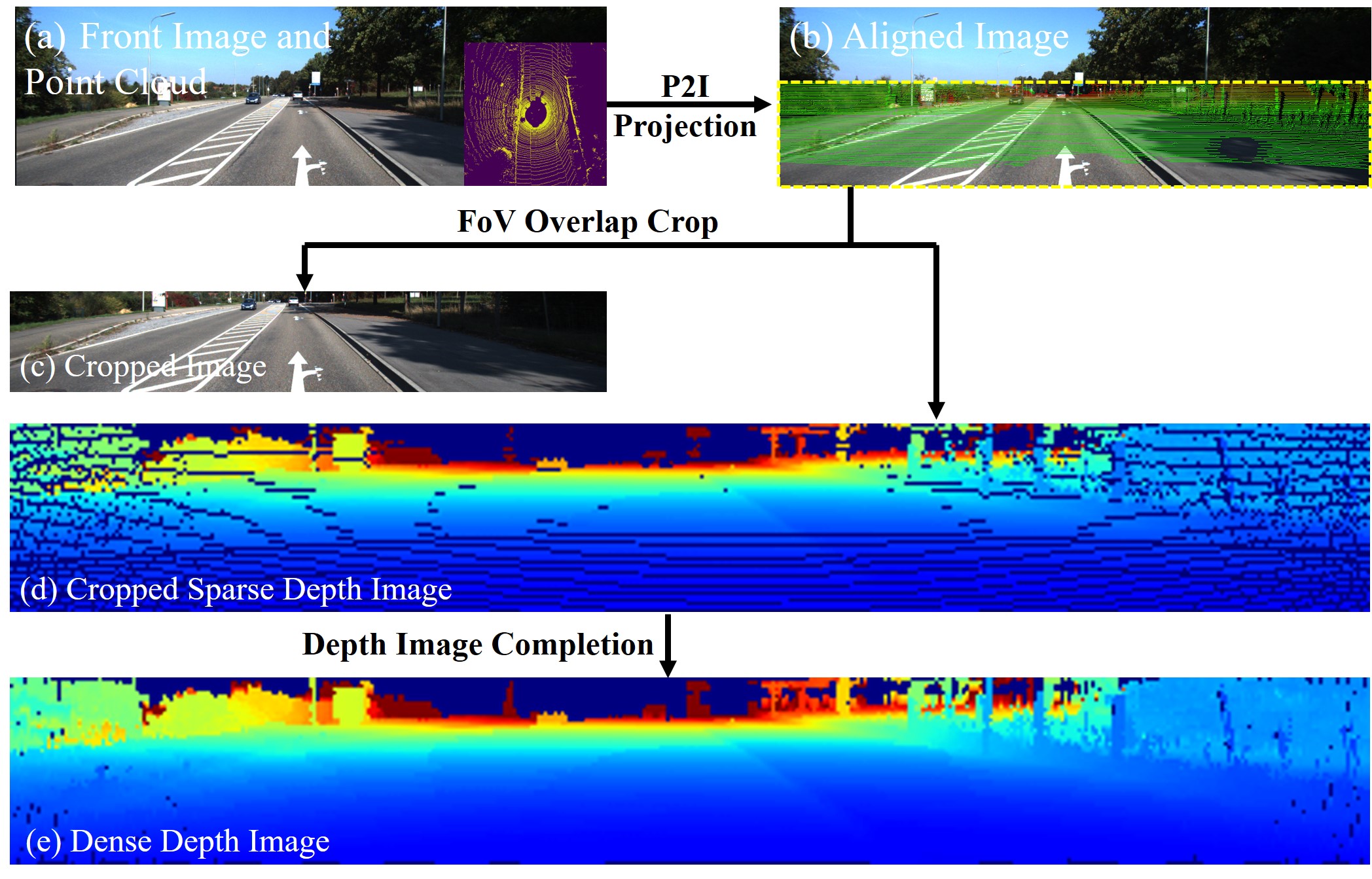}
	\caption{Visualization of image and point cloud alignment based on intrinsic and extrinsic matrix. After alignment, we crop the images based on FoV overlap and complete sparse depth images to generate dense depth images.}
	\label{align}
 \vspace{-1.0em}
\end{figure}

{\bf{Point Cloud to Image Projection.}}
We first project the point cloud $ P $ into depth image $ D $ to reduce the gap between different modalities. LiDAR point clouds are provided with Cartesian coordinates $ (x,y,z) $. According to the principles of pin-hole camera geometry, we project each point $ p(x,y,z) $ in the point cloud into the image plane with its coordinates $ (u, v) $ as
\begin{equation}
	[u, v]^{T} = \mathbf{K}(\mathbf{R}[x, y, z]^{T}+\mathbf{t})\label{projection},
\end{equation}
where $ \mathbf{K} $ is the camera's intrinsic matrix, and $ \mathbf{R} $ and $ \mathbf{t} $ correspond to the extrinsic rotation matrix and translation vector, respectively.
The corresponding pixel intensity $ D(u,v) $ of the depth image is calculated as the depth of point and can be formulated as $ D(u,v) = \sqrt[]{x^2 + y^2 + z^2} $.
If two points are projected onto the same pixel of the image, we choose the one with the smaller depth and discard the farther one.

{\bf{FoV Overlap Crop.}}
The sizes of the point cloud depth image and RGB image are usually different since LiDAR scanners and cameras have different FoV.
Typically, multi-beam LiDAR sensors have a 360-degree field of view, whereas the cameras are limited to perceiving a cone region in their frontal view. To deal with this, we crop the RGB images and depth images into a fixed-size window, the max elevation angle of which is limited to 5 degrees according to the FoV overlap of the two specific sensors.

\begin{figure}[!t]
	%\vspace{-0.5em}   %调整图表上方与正文之间的距离
	%\setlength{\abovecaptionskip}{-0.1cm}   %调整图片标题与图距离
	\setlength{\belowcaptionskip}{-0.3cm}
	\centering
	\includegraphics[width=0.5\textwidth]{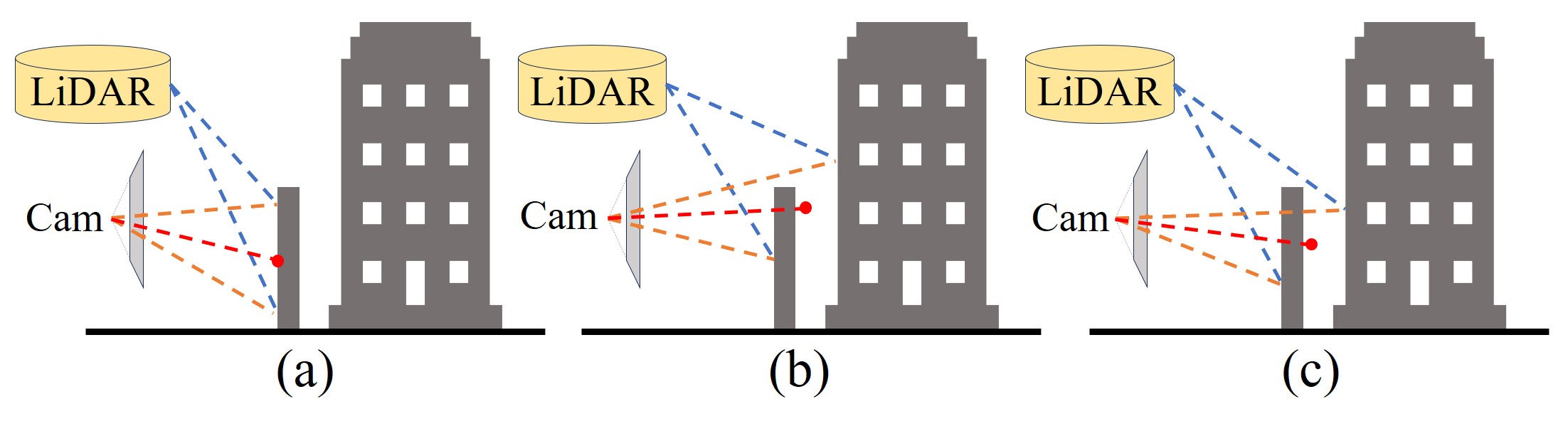}
	\caption{Three cases of depth image completion. The red point represents a pixel on the image plane where no LiDAR point is projected.}
	\label{inter}
\end{figure}

{\bf{Depth Image Completion.}}
In urban driving scenarios, most objects stand on the ground plane. In this situation, vertically adjacent points in point clouds are more likely to come from the same object \cite{liu2020globally}. We take advantage of this feature to generate interpolating points according to vertically adjacent pixels to achieve depth image completion. Fig. \ref{inter} provides a description of this completion procedure. As shown in Fig. \ref{inter}(a), the two blue lines are adjacent LiDAR scanning lines and the blue points are two points in the point cloud. We project these two points onto the depth image as two pixels. Then, there is a fissure pixel between the above two pixels obtained by projection. We interpolate the fissure pixel and back-project it into the 3D space to obtain the red points. Ideally, the red dot should be on the surface of the object.
Let $ D $ denote the depth image of the point cloud $ P $, and $ D(u', v') $ is a fissure pixel to which no LiDAR point corresponds. We interpolate the missing pixel $ D(u', v') $ with $ D(u'+i, v') $ and $ D(u'-j, v') $ through linear interpolating,
\begin{equation}
	D(u', v) = \dfrac{jD(u'+i, v') + iD(u'-j, v')}{i+j},
\end{equation}
where $ i,j>0 $.

\begin{figure}[!t]
	%\vspace{-0.5em}   %调整图表上方与正文之间的距离
	%\setlength{\abovecaptionskip}{-0.1cm}   %调整图片标题与图距离
	\setlength{\belowcaptionskip}{-0.5cm}
	\centering
	\includegraphics[width=0.5\textwidth]{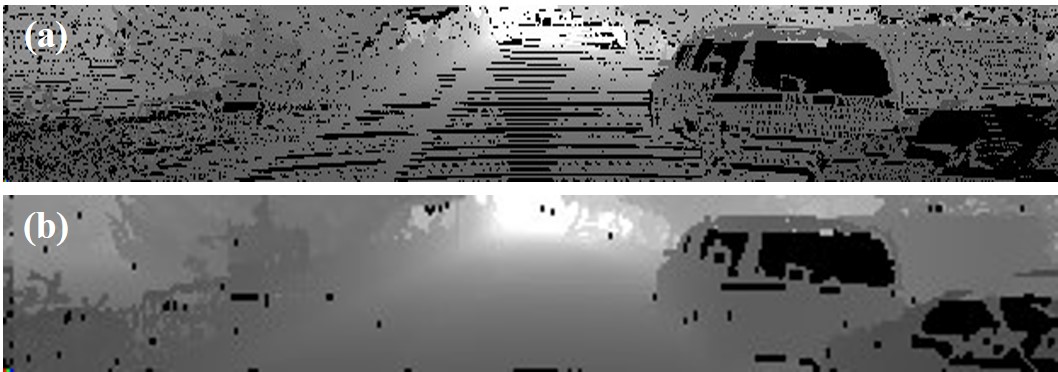}
	\caption{A demonstration of depth image interpolation.}
	\label{inter2}
\end{figure}

Noted that, due to the occlusion between the background and the foreground objects, two adjacent pixels in the depth image may be nonadjacent in the source point cloud. This phenomenon will cause the interpolation point (noted as the red point) to appear between the foreground and the background instead of actually being on the surface of the foreground object, leading to incorrect interpolation results, as shown in Fig. \ref{inter} (b)(c).
To solve this problem, we set a threshold $ \sigma $. If $ \vert D(u+i, v) - D(u-j, v) \vert > \sigma $, we regard the larger one as the background point, and to keep the foreground object's outline, we make $ D(u, v) = min(D(u+i, v), D(u-j, v)) $. Fig. \ref{inter2} demonstrates an example of the above-mentioned depth image completion method.

% \subsection{Descriptor Encoder}
% We design a shared weight descriptor encoder that encodes depth images and RGB images into distinctive global descriptors. Although the depth image and the RGB image that describe the same place express the contents of the scene in different ways, the semantics contained in both are the same. Through training, the designed encoder can extract deep semantic information from the depth image and RGB image of the same place, and then encode them into similar descriptors. The details of the encoder are described in the next section.

\subsection{Novel NMF-Based Encoder Network}
After obtaining similar LiDAR and camera data representations, we design a shared weight descriptor encoder to further encode depth images and RGB images into distinctive global descriptors for cross-modal place recognition. As illustrated in Fig. \ref{network}, we first employ a CNN to extract local features. We then use the Non-negative Matrix Factorization module to extract additional semantic features, mining potential semantic consistency between the image and its corresponding point cloud. Two NetVLAD networks aggregate local and semantic features and generate global descriptors. Each module is detailed as follows:

\begin{figure}[!t]
	%\vspace{-0.5em}   %调整图表上方与正文之间的距离
	%\setlength{\abovecaptionskip}{-0.1cm}   %调整图片标题与图距离    
	%\setlength{\belowcaptionskip}{-0.5cm}
	\centering
	\includegraphics[width=0.48\textwidth]{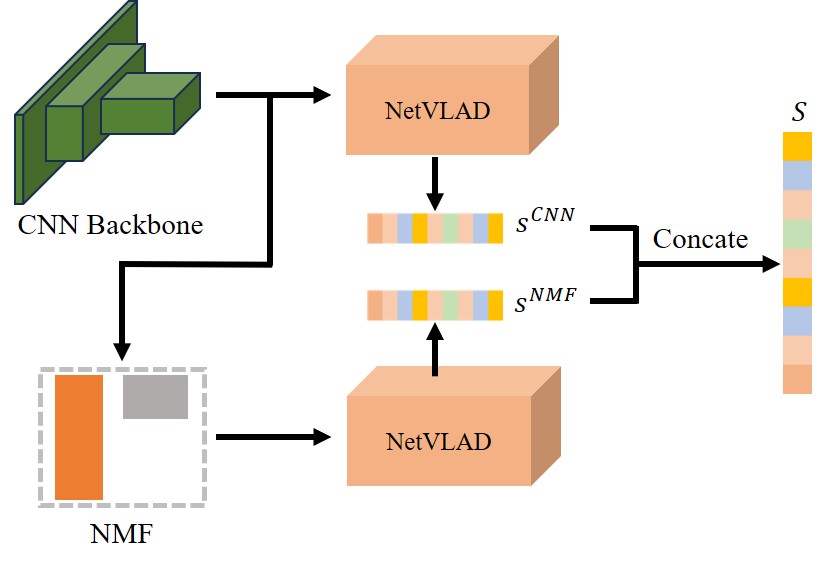}
	\caption{The structure of the shared-weight global descriptor encoder. It contains a CNN backbone, an NMF module and two NetVLAD networks.}
	\label{network}
\end{figure}

{\bf{Local Feature Extraction.}}
For an RGB or depth image input $ x $, we first extract local feature map $ f(x) \in \mathbb{R}^{H \times W \times C} $ using a CNN. In this representation, $ H $ and $ W $ denote the size of the output from the last CNN layer, and $ C $ represents the dimension of the feature space. We use ResNet34~\cite{he2016deep} as our local feature extractor.

{\bf{NMF Semantic Feature Extraction.}}
Semantic information is theoretically helpful to determine whether two observations are apparently similar through high-level scene understanding~\cite{dube2017segmatch}. However, the use of semantic segmentation networks can be computationally inefficient and require large amounts of manually annotated data for training. Employing such techniques may lead to slow inference speeds and introduce errors in semantic estimation, which can negatively affect place recognition. In ModaLink, we introduce a lightweight Non-negative Matrix Factorization (NMF) based encoder. This module has inherent clustering property~\cite{ding2005equivalence} and can mine latent semantic information in an unsupervised way. By exploiting such implicit semantic information, our NMF-based encoder can produce more distinctive descriptors while requiring no semantic labels.

NMF is a matrix decomposition method. It factorizes a non-negative feature matrix $ A $ into a non-negative cluster matrix $ P $ and a non-negative orthogonal matrix $ Q $. Here non-negative means that the matrices have no negative elements. NMF achieves an approximation of $ A \approx PQ $ by minimizing the error matrix, i.e.
\begin{equation}
	\min\Vert A-PQ \Vert _F, P\ge0, Q\ge0,
\end{equation}
where $\Vert\cdot\Vert_F$ means the Frobenius Norm. By imposing an orthogonality constraint on~$ Q $, where~$ QQ^T = I $, the aforementioned minimization is equivalent to K-means mathematically. Each row of $ Q $ can be interpreted as a cluster centroid in $ C $ dimensions due to the devised orthonormal constraints of NMF. Each row of the matrix P corresponds to the spatial position of a pixel from the matrix A. It represents the pixel feature as a particular cluster. NMF has a built-in clustering property that automatically groups the columns of matrix $A$ into clusters~\cite{ding2005equivalence}. Each row of the matrix $Q$ can be viewed as a cluster centroid of C dimensions, which corresponds to a coherent object among views. As a result, the cluster matrix $P$ can serve as a feature vector.

% The 
% The matrix $P$ represents the degree of association of each observation with a particular cluster. Each row of $P$ corresponds to a cluster and signifies the input's affinity to that cluster. NMF has a built-in clustering property that automatically groups the columns of matrix $A$ into clusters \cite{ding2005equivalence}. As a result, the cluster matrix $P$ can serve as a feature vector.

\begin{figure}[!t]
	%\vspace{-0.5em}   %调整图表上方与正文之间的距离
	%\setlength{\abovecaptionskip}{-0.1cm}   %调整图片标题与图距离    
	%\setlength{\belowcaptionskip}{-0.5cm}
	\centering
	\includegraphics[width=0.45\textwidth]{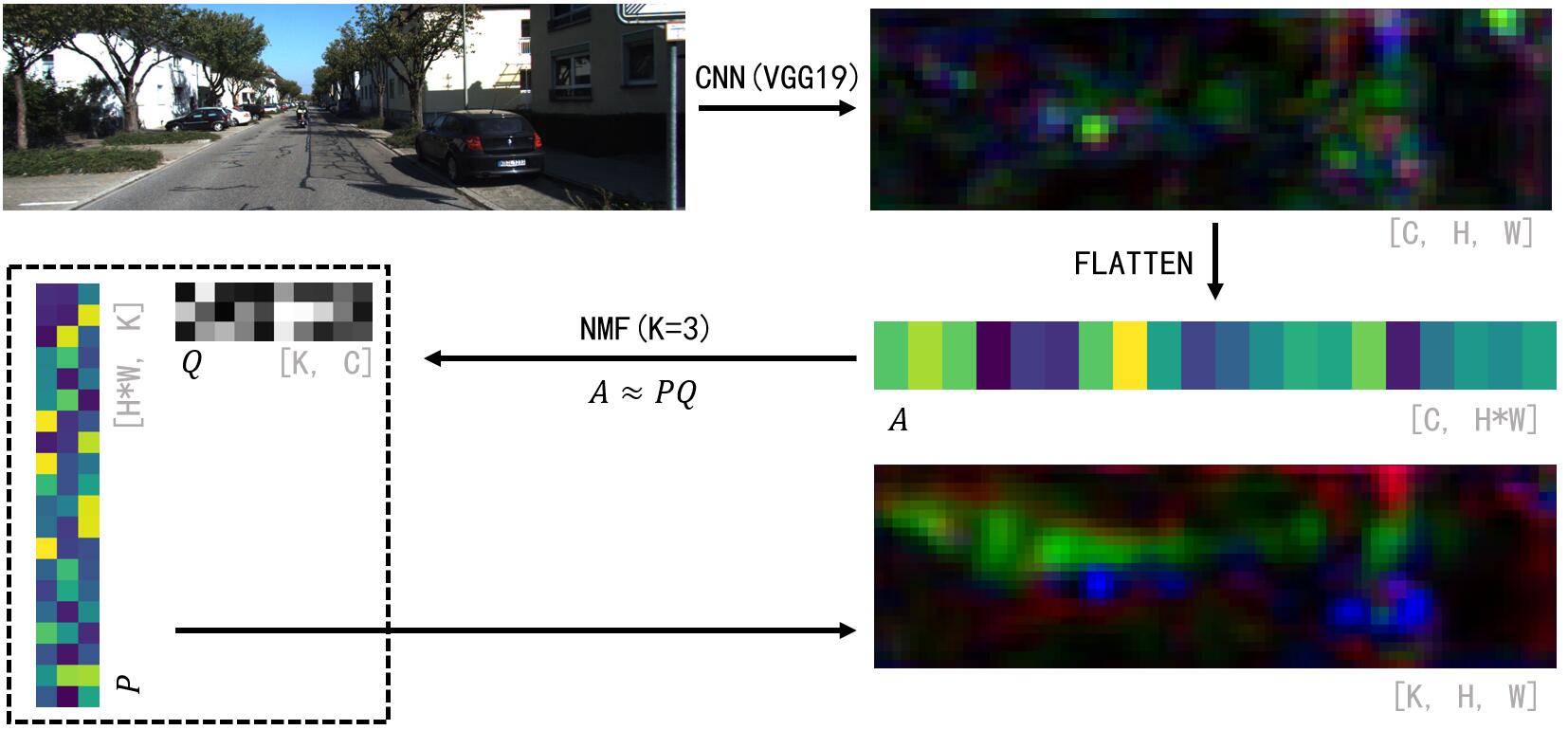}
	\caption{The process of the NMF module.}
	\label{nmf1}
\end{figure}

\begin{figure}[!t]
	\vspace{-0.5em}   %调整图表上方与正文之间的距离
	\setlength{\belowcaptionskip}{-0.3cm}
	\centering
	\includegraphics[width=0.45\textwidth]{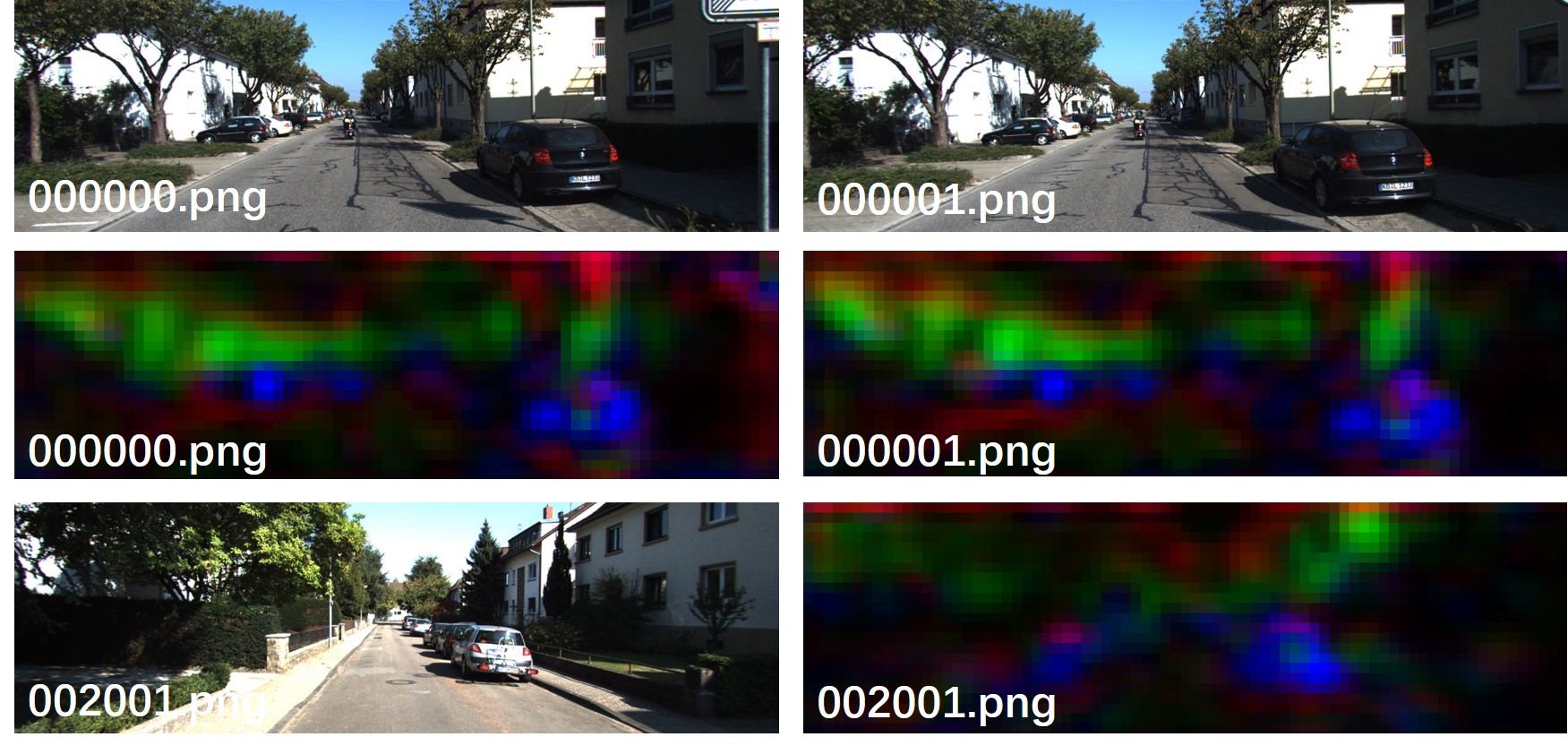}
	\caption{Brief illustration of the clustering effect of NMF.}
	\label{nmf2}
\end{figure}

% \noindent {\bf{NMF Features.}}
Since the feature map activated by ReLU inherently consists of non-negative elements, we use NMF to factorize the feature map $A$ to generate explicit semantic features.  Fig.~\ref{nmf1} shows the factorization procedure. 
The extracted feature map from the local feature extraction module has dimensions ($ H, W, C $), and these feature maps are concatenated and reshaped into an ($ NHW, C $) matrix $ A $. Each row of $ A $ represents a feature vector of an observation. By solving NMF \cite{ding2005equivalence}, the $ A $ matrix is factorized into a ($ NHW, K $) matrix $ P $ and a ($ K, C $) matrix $ Q $, where $ K $ represents the number of semantic clusters, and $ A \approx PQ $.  $ P $ captures the similarity between each pixel and the clustered semantic objects, and are then employed as our semantic features.

Fig. \ref{nmf2} illustrates examples of extracted semantic feature maps with our proposed encoder network, where the NMF exhibits a semantic clustering property. For similar input RGB images, NMF can generate similar semantic feature maps. For dissimilar RGB images, NMF can extract similar features from the parts of the image with the same semantics. 

{\bf{Global descriptor generation.}}
To generate global place descriptors, we finally leverage NetVLAD~\cite{arandjelovic2016netvlad} to cluster local features.
The local feature map $ f(x) $ serves as the input of NetVLAD and is then transformed into a partial global localization descriptor $ s^{CNN} \in \mathbb{R}^{D_{CNN}} $, where $ D_{CNN} $ is the dimension of the descriptor. Following the NMF module, $ f(x) $ is clustered into $ g(x) \in \mathbb{R}^{H \times W \times K} $. Another NetVLAD transformation is applied to $ g(x) $, resulting in $ s^{NMF} \in \mathbb{R}^{D_{NMF}} $. Finally, the concatenation of $ s^{CNN} $ and $ s^{NMF} $ forms the ultimate global place descriptor $ s $.

\subsection{Training And Inference}
Before training, we construct a keyframe database to represent a global point cloud map. We find positive and negative matches for the query image according to ground truth distance. Because in the subsequent experiments, when the positioning result is within 10 meters of the ground truth, we regard it as a correct positioning. So, in the training process, we regard the point clouds within the stricter 5-meter range as the point cloud located in the same place, and accordingly, for each query image, we regard the point clouds within 5 meters away from the query as positive samples, while other point clouds as negative samples.

For supervision, we adopt lazy-triplet loss \cite{schroff2015facenet} to guide the model encoding the camera and depth images into descriptors that map the same location close in the feature space, while different locations farther in the feature space:
\begin{equation}
	\mathcal{L}_{triplet}=\sum_i^n max(m + d(s_q, s_p) - d(s_q, s_n^i), 0),
\end{equation}
where $ m $ is the constant margin set as 0.3. $ s_q $, $ s_p $, and $ s_n^i $ denote the learned descriptors from the query image, positive depth image, and the $ I $-th negative depth image. The distance between descriptors is defined by $ d(x,y) = \Vert x - y \Vert_2 $.

The global pose information of the prior database can be obtained using a SLAM system or RTK information. We sample LiDAR keyframes and generate their global descriptors using our Modalink at intervals of 5 meters.
During the inference phase, we use NMF-Encoder to extract the global descriptor for the incoming query image and employ the Faiss approach~\cite{johnson2019billion} for fast retrieval of the best-matched LiDAR frame. The global pose of the query image is then determined according to the corresponding position of the matched LiDAR frame.

\section{Experiments}

In this section, we evaluate the performance of our ModaLink for cross-modal on point cloud maps place recognition. Based on the KITTI \cite{kitti} and our self-collected dataset, we evaluate the performance of ModaLink and draw comparisons with baseline methods \cite{zheng2023i2p, cattaneo2020global}. Finally, we conduct ablation studies to analyze crucial components and assess the runtime requirements of our method.

\subsection{Datasets}

% \begin{table}[t]
% 	\renewcommand\arraystretch{1.1}
%         \setlength{\tabcolsep}{5pt}
% 	\caption{Partition of the dataset.}
% 	\begin{center}
% 		\begin{tabular}{ccccccc}
% 			\toprule
% 			\textbf{\ } & Train  & Val 		 & \multicolumn{4}{c}{Test} 		  \\
% 			\midrule
% 			sequence	& 00	 & 00		 & 02	  & 05		& 06	 & 08 	  \\
% 			frame		& 0-3000 & 3200-4540 & 0-4660 & 0-2760	& 0-1100 & 0-4070 \\
% 			\bottomrule
% 		\end{tabular}
% 	\end{center}
% 	\label{kitti}
%  \vspace{-2.0em}
% \end{table}

{\bf{KITTI.}}
The KITTI contains a large number of synchronized and calibrated point clouds and images. It is widely used to verify autonomous driving-related algorithms. We use the images and point clouds of 0-3000 frames in sequence 00 for training and the rest of the frames in sequence 00 for validation. We use the 02, 05, 06, and 08 sequences of its odometry subset for testing.

\begin{figure}[!t]
	\setlength{\belowcaptionskip}{-0.3cm}
	\centering
	\includegraphics[width=0.45\textwidth]{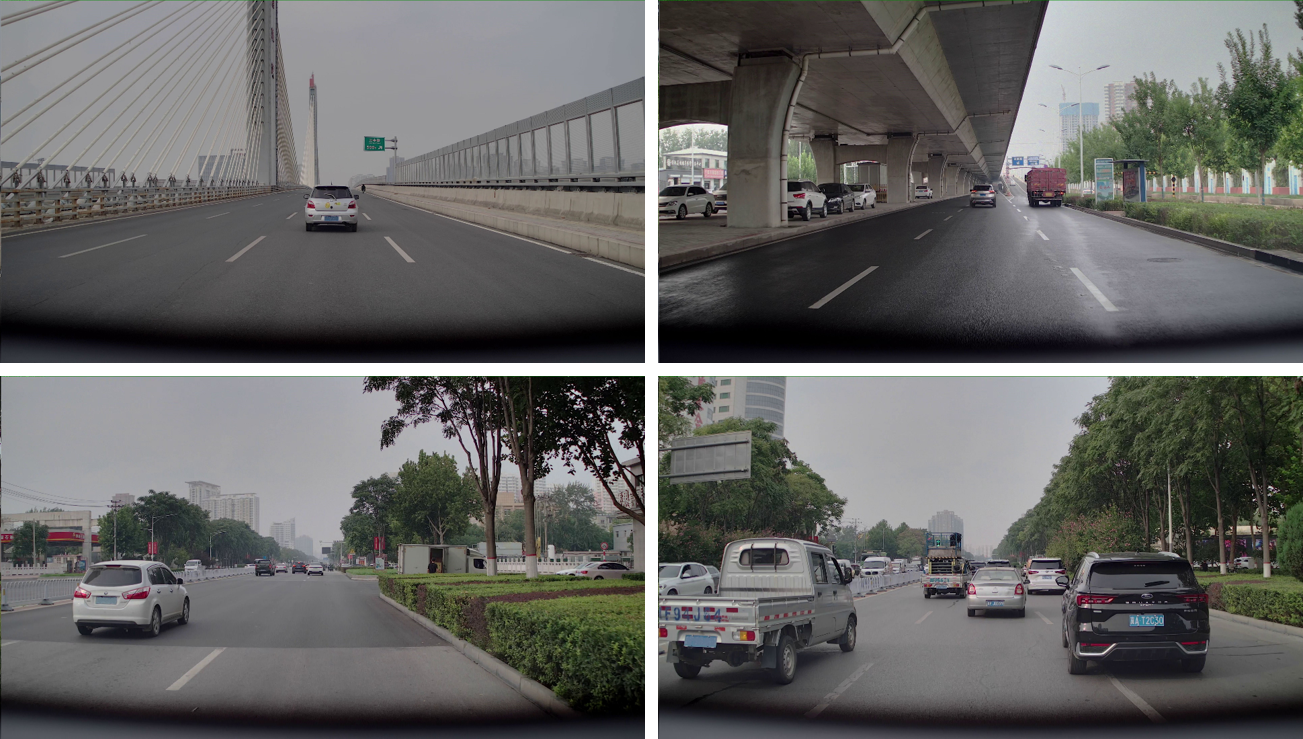}
	\caption{Some sample images from HAOMO Dataset.}
	\label{haomodataexample}
\end{figure}

{\bf{HAOMO Dataset.}}
To assess the generalization capability of ModaLink, we further use the HAOMO Dataset~\cite{OverlapTransformer}. This dataset was collected by the HAOMO.AI company in China, with a vehicle equipped with a LiDAR sensor (128 beams), a wide-angle camera, and an RTK GNSS. The HAOMO dataset encompasses a trajectory length of over 17.4 km. The scenes include an overpass and a town. It comprises 25,278 pairs of point clouds and images. Fig.~\ref{haomodataexample} displays some sample images from HAOMO dataset. 

\subsection{Evaluation Metrics}
For evaluation, we adopt the recall at Top-N and the recall at Top-1\% as metrics. When performing the matching, we extract global descriptors of query images and retrieve Top-N nearest matches from the point cloud database. Following the setup of I2P-Rec~\cite{zheng2023i2p}, We regard a match between a query image and the point cloud as true positive when their geometric distance is less than a threshold of $ t=10 $ meters. Recall at Top-N is computed as the proportion of correct predictions among all positive instances in a dataset considering the Top-N prediction for each sample.

\subsection{Implementation Setup}

Our ModaLink employs a pretrained ResNet34 as the backbone, with the number of NMF clusters $ K=16 $. NetVLAD modules have a cluster number of $ d_{k}=64 $. The intermediate feature channels of the ResNet and NMF branches are set to 256 and $ K $, respectively. The output descriptor of ModaLink is a float vector of length $ 256 \times 64 + K \times 64 $. 
We use data augmentation to enhance the model's robustness to variations. Specifically, we incorporate random rotation within $ \pm 5^{\circ} $ and position shifts within $ \pm 0.1m $ to the point cloud before projection. More implementation details can be found at our open-source repository\footnote{https://github.com/haomo-ai/ModaLink.git}.

\begin{table*}[!t]
	\setlength{\abovecaptionskip}{-0.1cm}
	\caption{The recall@1 and recall @1\% of place recognition methods on the KITTI dataset.}
	\begin{center}
		\begin{tabular*}{0.95\textwidth}{@{\extracolsep{\fill}} cccccccccccc}
			\toprule
			Sequence		& \multicolumn{2}{c}{00} & \multicolumn{2}{c}{02} & \multicolumn{2}{c}{05} & \multicolumn{2}{c}{06}	& \multicolumn{2}{c}{08} \\
			Recall			& @1 & @1\% & @1 & @1\% & @1 & @1\% & @1 & @1\% & @1 & @1\% \\
			\midrule
			Baseline 	& 58.0 & -		& 4.0  	& -		& 10.6 & -		& 22.3 & -		& 5.3 	& -	  \\
			MIM-Points 	& 53.3 & 71.6 	& 6.9 	& 57.1 	& 14.8 & 68.3 	& 20.0 & 44.4 	& 12.5 	& 68.6 \\
			PSM-Points*	& 58.0 & 72.3	& 8.2	& 53.6 	& 16.0 & 71.1 	& 32.1 & 50.0	& 14.7	& 75.0 \\
			LEA-Points*	& 58.8 & 72.6 	& 9.1 	& 61.3 	& 21.8 & 80.0 	& 31.6 & 71.6 	& 18.0 	& 76.3 \\
			MIM-I2P-Rec	& 74.3 & 88.6	& 46.3 	& 89.7	& 49.9 & 89.1	& 41.9 & 81.5	& 42.1 	& 88.4 \\
			PSM-I2P-Rec*& 81.7 & 97.4	& 47.6 	& 92.9	& 63.8 & 94.1	& 52.1 & 88.5	& 53.5 	& 92.1 \\
			LEA-I2P-Rec*& 92.6 & 99.7	& \bf{77.0} & \bf{98.4}	& 83.4 & 98.8 & 55.5 & 91.5	& 69.4 & 96.4 \\
			Ours	& \bf{98.0} & \bf{100}	& 70.5 & 97.7 & \bf{91.3} & \bf{99.3} & \bf{87.4} & \bf{100} & \bf{84.4} & \bf{99.9} \\
			\bottomrule
		\end{tabular*}
	\end{center}
	\label{kittirecall}
	\vspace{-0.5em}
    \captionsetup{justification=raggedright, singlelinecheck=false}
	\caption*{\textit{\ \ \ * indicates that these methods require stereo images as input.}}
	\vspace{-0.5cm}
\end{table*}

\subsection{Performance on the KITTI}
We assess the place recognition performance on the KITTI dataset, and compare our method with baseline \cite{cattaneo2020global} and I2P-Rec \cite{zheng2023i2p}. 
The baseline method uses 2D CNN and 3D DNN to process 2D images and 3D point clouds separately, and then generate descriptors with NetVLAD~\cite{arandjelovic2016netvlad}.
I2P-Rec uses monocular depth estimation, i.e. MIM~\cite{xie2023darkmim} or stereo matching, i.e. PSM~\cite{chang2018psm} and LEA~\cite{cheng2020lea} to recover point clouds from RGB images, and generates descriptors based on the BEV images of clouds. For a more comprehensive evaluation, in I2P-Rec, Zheng et al.~\cite{zheng2023i2p} also extract global descriptors from the recovered clouds with PointNetVLAD~\cite{angelina2018pointnetvlad}.
They denote this method using different depth estimation algorithms as MIM-Points, PSM-Points, and LEA-Points.

\begin{figure*}[!t]
	%\vspace{-0.5em}   %调整图表上方与正文之间的距离
	%\setlength{\abovecaptionskip}{-0.1cm}   %调整图片标题与图距离    
	%\setlength{\belowcaptionskip}{-0.5cm}
	\centering
	\includegraphics[width=\textwidth]{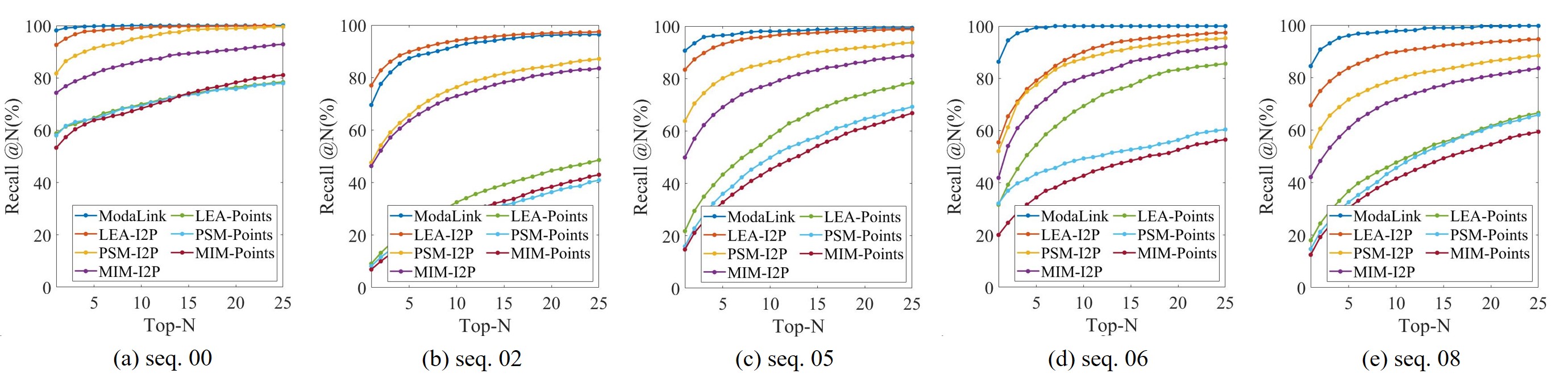}
        \vspace{-0.5cm}
	\caption{The Top-N recall rate on the KITTI dataset.}
	\label{topn}
	\vspace{-1.0em}
\end{figure*}

As shown in Table \ref{kittirecall}, our proposed ModaLink exhibits a significant improvement, outperforming stereo-image-based and depth-estimation-based I2P-Rec in most scenarios. Fig. \ref{topn} visually depicts the recall rates at Top-N, where ModaLink consistently outperforms other methods. Note that our method only uses monocular images as queries while outperforming methods exploiting stereo images.

\subsection{Performance on the HAOMO dataset}

\begin{table}[t]
	\setlength{\abovecaptionskip}{-0.1cm}
	\caption{Generalization on the HAOMO dataset.}
	\begin{center}
		\begin{tabular*}{0.4\textwidth}{@{\extracolsep{\fill}} cccc}
			\toprule
			 & Recall@1		& Recall@1\%	\\
			\midrule
			Trained on KITTI	& 35.5		& 75.5	\\
			Refined	on HAOMO	& 70.9		& 98.2	\\
			\bottomrule
		\end{tabular*}
	\end{center}
	\label{hmd}
 \vspace{-0.4cm}
\end{table}
\begin{table}[t]
	\setlength{\abovecaptionskip}{-0.1cm}
	\caption{Performance of ModaLink against different thresholds on HAOMO dataset.}
 \setlength{\tabcolsep}{3pt}
	\begin{center}
		\begin{tabular*}{0.49\textwidth}{@{\extracolsep{\fill}} ccccccccccc}
			\toprule
                Threshold(m) & 5  & 10  & 15  & 20  & 25  & 30  & 35  & 40  & 45  & 50 \\
			\midrule
			Recall@1 & 69.1&70.9&71.8&72.6&73.3&73.6&73.9&74.2&74.6&74.9 \\
			\bottomrule
		\end{tabular*}
	\end{center}
	\label{haomotop1t}
 \vspace{-0.5cm}
\end{table}

HAOMO dataset was collected along busy streets with significant interference and utilized this dataset for evaluation. To validate the generalization ability of our method, we designate 30\% of the LiDAR frames and corresponding images for training, reserving the remainder for testing. As shown in Table.~\ref{hmd}, ModaLink achieves a Top-1 recall rate of 35.5\% while only trained on the KITTI and a Top-1 recall rate of 70.9\% while refined on the HAOMO dataset, validating its generalization ability. As shown in Table~\ref{haomotop1t}, after further training on 30\% of the HAOMO dataset, ModaLink performs satisfactorily under different positive thresholds.

\begin{figure}[!t]
	%\vspace{-1em}   %调整图表上方与正文之间的距离
	%\setlength{\abovecaptionskip}{-0.1cm}   %调整图片标题与图距离    
	%\setlength{\belowcaptionskip}{-0.5cm}
	\centering
	\includegraphics[width=0.5\textwidth]{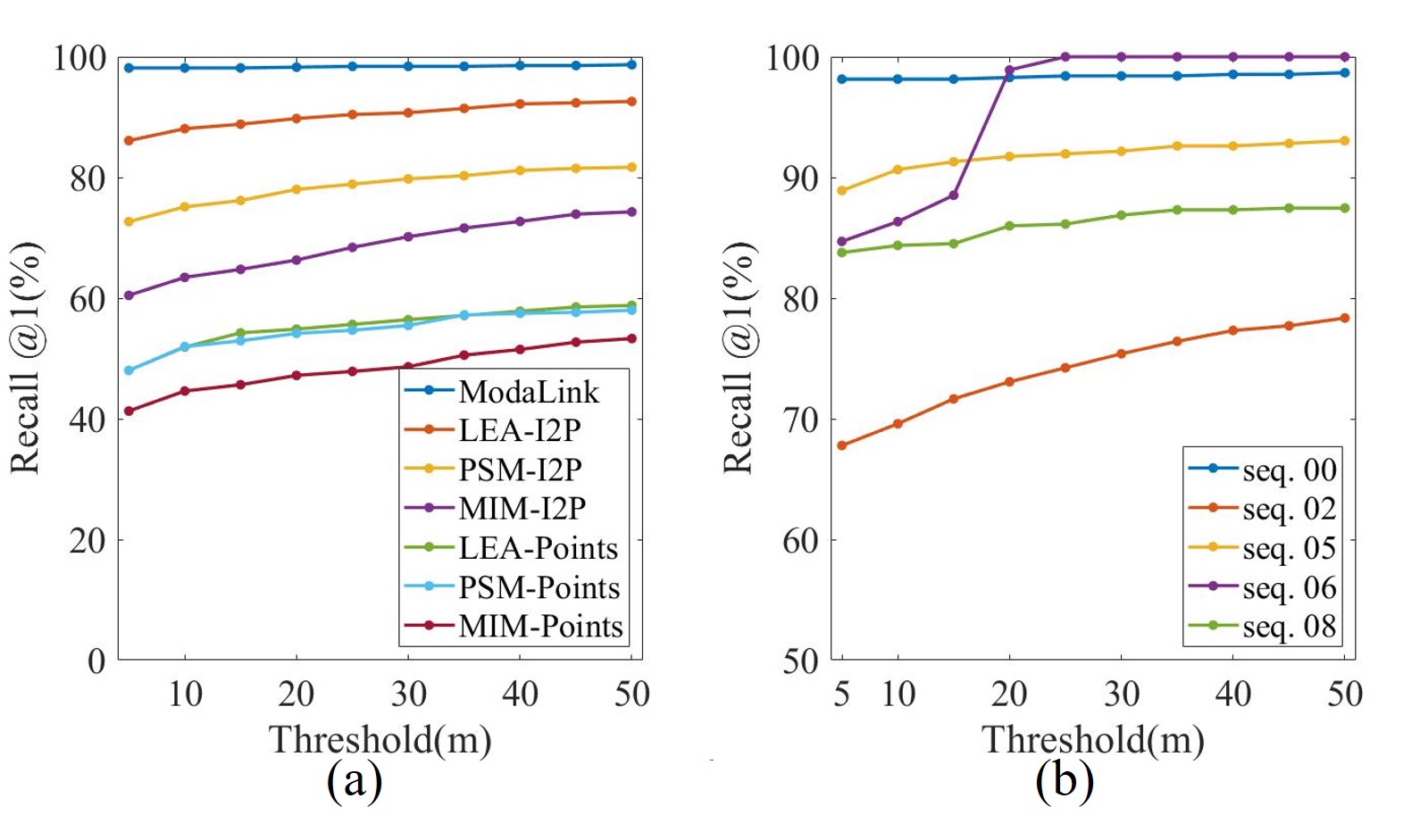}
	\caption{The Top-1 recall with respect to different threshold on seq 00 and the Top-1 recall with respect to different sequence and threshold for ModaLink.}
	\label{toptandtopt5}
 \vspace{-1.0em}
\end{figure}

Additionally, we explore the sensitivity of different methods against the ground truth threshold $ t $ on both datasets. As depicted in Fig. \ref{toptandtopt5}(a), recall@1 increases as $ t $ varies from 5m to 50m, with our ModaLink maintaining a consistently high recall@1. As shown in Fig. \ref{toptandtopt5}(b), ModaLink also has a satisfactory performance on other sequences which further approves its generalization ability.

\subsection{Ablation Study}

We conduct detailed ablation studies for the depth completion (DC) in FoV Transformation Module and the NMF on the KITTI dataset.

\begin{table}[!t]
	\vspace{0.5em}
	\setlength{\abovecaptionskip}{-0.1cm}
	\caption{Ablation study of different components in our ModaLink.}
	\begin{center}
		\begin{tabular*}{0.45\textwidth}{@{\extracolsep{\fill}} cc|cccccc}
			\toprule
			NMF			& DC		& 00		& 02		& 05		& 06		& 08	\\
			\midrule
			& 			& 26.1 		& 12.9 		& 32.8 		& 64.5		& 22.1
			\\
			\checkmark	& 			& 83.4		& 69.5		& 83.4		& 55.5		& 69.4	\\
			& \checkmark& 93.1		& 62.8		& 88.3		& 85.8		& 77.1
			\\
			\checkmark	& \checkmark& 98.0		& 70.5		& 91.3		& 87.4		& 84.4
			\\
			\bottomrule
		\end{tabular*}
	\end{center}
	\label{ablation}
	\vspace{-1.5em}
\end{table}

{\bf{Depth Completion in FoV Transformation Module.}}
We compare our ModaLink with the variants of not using depth completion in the FoV transformation module. Comparing four different setups, as shown in Table. \ref{ablation}, ModaLink with DC outperforms the setups without DC for place recognition. The results show that dense depth images can help generate more distinctive place descriptors.

{\bf{NMF Segmentation Module.}}
As shown in Table. \ref{ablation}, NMF+DC outperforms the other setups for place recognition. The results show that our NMF module increased the performance significantly, and potential semantic information lead to higher efficiency.

\begin{table}[t]
	\setlength{\abovecaptionskip}{-0.1cm}
	\caption{Comparation of different numbers of semantic clusters $ K $.}
	\begin{center}
		\begin{tabular*}{0.45\textwidth}{@{\extracolsep{\fill}} c|cccccc}
			\toprule
			Clusters& 00		& 02		& 05		& 06		& 08		\\
			\midrule
			3		& 89.6		& 38.1		& 75.7		& 76.5		& 60.8		\\
			5		& 93.5		& 60.8		& 87.4		& 85.3		& 73.9		\\
			16		& 98.0		& 70.5		& 91.3		& 87.4		& 84.4		\\
			64		& 94.1		& 63.7		& 86.7		& 84.7		& 77.1		\\
			\bottomrule
		\end{tabular*}
	\end{center}
	\label{nmfk}
 \vspace{-0.5cm}
\end{table}

{\bf{NMF Cluster Number.}}
The number of clusters, denoted as $ K $, is a crucial hyperparameter for NMF. To explore its effect, we conducted experiments with different cluster numbers. The results, as presented in Table \ref{nmfk}, indicate that a higher number of clusters does not consistently result in superior performance. An ideal $ K $ is observed to be around 16 for ModaLink.

\subsection{Runtime}
This experiment evaluates the runtime requirements of our method. All experiments are conducted on a computer equipped with an Intel Xeon Platinum 8362 CPU and an Nvidia A100 SXM4 GPU. 

The NMF-Encoder takes an average of 22.51ms to encode an image into a descriptor. The descriptor retrieval takes averages 10.94ms. Overall, our ModalLink takes 33.45ms to localize an image and achieve an FPS of about 30Hz In contrast, the depth estimation part of I2P-Rec, namely LEA-Stereo, consumes at least 0.3s to estimate the depth for a pair of stereo images. This showcases the evident advantage in the speed of ModaLink.

\section{Conclusion}

We presented ModaLink, a novel framework for cross-modal place recognition,  specifically addressing Image-to-PointCloud tasks in autonomous vehicles. Our approach introduces a Field of View (FoV) transformation module for real-time performance and eliminates the need for computationally intensive depth estimation modules. We also design a Non-negative Matrix Factorization (NMF) based module to improve the distinctiveness of descriptors.
Experimental results on the KITTI dataset showcase the state-of-the-art performance of ModaLink. Extensive evaluations on the HAOMO dataset further validate the system's practical generalization capabilities in diverse environments. The open-source implementation of our methods is available, contributing a fast solution to cross-modal place recognition for autonomous vehicles.

\bibliographystyle{ieeetr} 
\bibliography{references}

\end{document}